\documentclass[conference, 10pt]{IEEEtran}
\IEEEoverridecommandlockouts
\usepackage{cite}
\usepackage{amsmath,amssymb,amsfonts}
\usepackage{algorithmic}
\usepackage{graphicx}
\usepackage{textcomp}
\usepackage{xcolor}
\usepackage{subfigure}
\usepackage{makecell}
\usepackage{multirow}
\usepackage{todonotes}
\usepackage{tcolorbox}
\usepackage{diagbox}
\usepackage{array}
\usepackage{verbatim}
\def\BibTeX{{\rm B\kern-.05em{\sc i\kern-.025em b}\kern-.08em
    T\kern-.1667em\lower.7ex\hbox{E}\kern-.125emX}}

\makeatletter
\renewcommand{\citepunct}{,\penalty\@m\hskip.13emplus.1emminus.1em}
\renewcommand{\citedash}{\hbox{--}\penalty\@m}

\begin{document}

\title{Learn to Optimize Resource Allocation under QoS Constraint of AR
%{\footnotesize \textsuperscript{*}Note: Sub-titles are not captured in Xplore and should not be used}
%\thanks{Identify applicable funding agency here. If not, delete this.}
}

\author{
\thanks{This work was supported in part by the National Natural Science Foundation of China (NSFC) under Grant 62271024 and in part by the National Key Research and Development Program of China under Grant 2022YFB2902002.}
\IEEEauthorblockN{Shiyong Chen, Yuwei Dai, Shengqian Han}
\IEEEauthorblockA{School of Electronics and Information Engineering, Beihang University, Beijing 100191, China\\Email: \{shiyongchen, ywdai, sqhan\}@buaa.edu.cn}
}

\maketitle

\begin{abstract}
This paper studies the uplink and downlink power allocation for interactive augmented reality (AR) services, where the live video captured by an AR device is uploaded to the network edge, and then the augmented video is subsequently downloaded. By modeling the AR transmission process as a tandem queuing system, we derive an upper bound for the probabilistic quality of service (QoS) requirement concerning end-to-end latency and reliability. The resource allocation under the QoS requirement results in a functional optimization problem. To address it, we design a deep neural network to learn the power allocation policy, leveraging the optimal power allocation structure to enhance learning performance. Simulation results demonstrate that the proposed method effectively reduces transmit power while meeting the QoS requirement.
\end{abstract}

\begin{IEEEkeywords}
Augmented reality (AR), end-to-end latency, power allocation, QoS.
\end{IEEEkeywords}

\addtolength{\topmargin}{-.6cm}
\section{Introduction}
Deploying augmented reality (AR) over wireless networks is a crucial step towards realizing the Metaverse~\cite{metaverse}. AR integrates virtual objects into a live view of the real world, creating a realistic and personalized interactive environment. To achieve a seamless, immersive wireless AR experience, stringent quality-of-service (QoS) requirements concerning end-to-end (E2E) latency and reliability should be satisfied, meanwhile, high data rates are required.

For AR services, the E2E latency requirement is modeled as the packet delay budget (PDB), where the delay budget defines the maximum allowable delay from the instant a live video frame is generated to the instant the corresponding augmented video frame is returned. Reliability can be modeled by the packet loss rate (PLR), which includes both the probability of packet transmission error and the probability that the E2E delay exceeds the PDB, known as the PDB violation probability~\cite{Standard}. In the mobile edge computing (MEC)-based wireless AR system, resource allocation to ensure latency and reliability requirements was investigated in \cite{Resource_allocation, QoS_Aware, Task_Offloading}. A federated learning approach was proposed in \cite{Resource_allocation} to minimize resource usage, where the PDB was treated as a hard constraint to ensure that the E2E latency of any packet does not exceed the PDB. In  \cite{QoS_Aware, Task_Offloading}, the total E2E delay was minimized under resource constraints. However, due to the fluctuation of wireless channels, using PDB as a hard constraint can result in unbounded resource utilization under poor channel conditions. Moreover, minimizing E2E delay may excessively satisfy the E2E latency requirement, leading to resource waste. By taking PDB as the latency constraint and allowing rare PDB violations, the efficiency of resource utilization can be significantly improved \cite{QoS_Guaranteed}.

Given the interactive nature of AR services, which involve uplink (UL) transmission, edge computing, and downlink (DL) transmission, an AR system can be modeled as a tandem queuing system. The PDB violation probability in such queuing systems was studied in \cite{EBEC, uVR, Two_side}. In \cite{EBEC}, the PDB violation probability for ultra-reliable and low-latency communication (URLLC) services was derived by effective capacity theory. However, this study considered short-packet data that can be transmitted within a coherent block, which does not apply to AR systems that require large packet sizes and high data rates. Stochastic network calculus (SNC) offers a versatile mathematical framework, capable of providing upper bounds of PDB violation probability by transforming complex queuing systems into analytically manageable linear models formulated in the min-plus (or max-plus) algebra framework\cite{uVR, Two_side}. In \cite{uVR}, SNC was employed to derive an upper bound on the probability of PDB violation for a VR system, with the DL transmission process being analyzed under the assumptions of hyper-exponential computation time and constant UL delay.

In this paper, we optimize the transmit power for both UL and DL in an interactive AR system while meeting the QoS constraints. To this end, we first derive an upper bound on the PDB violation probability with SNC and martingale theory by modeling the AR system as a tandem queuing system in the time domain. Next, we develop an approximate distribution for the service time of each packet, which establishes the relationship between QoS constraints and transmit power. The resulting bidirectional power allocation is a functional optimization problem, which poses challenges for traditional optimization methods. To address this, we design a deep neural network (DNN) to learn the power allocation policy, where the water-filling structure is exploited to enhance learning performance. Simulation results demonstrate that the proposed method achieves considerable power savings while satisfying the QoS requirements.

\section{System Model}
Consider a MEC-assisted wireless AR system, where an MEC-enabled base station (BS) equipped with $N_{t}$ antennas serves a single-antenna AR user. 
Due to the limited computation and power resources of the AR devices, the computation tasks, such as object detection and rendering, are offloaded to the MEC. This requires the AR user to upload the live video to the BS via UL transmission. The MEC then detects target objects in the received video, generates virtual objects, and superimposes them onto the detected objects. The augmented video is compressed and sent to the AR device via DL transmission. Since medium-quality UL video is sufficient for the detection of target objects, the UL video stream is often downscaled to reduce transmission requirements compared to the higher-quality DL stream \cite{Standard}.

\subsection{QoS Requirement of AR System}
\begin{figure}%[htbp]
\centerline{\includegraphics[width=0.45\textwidth]{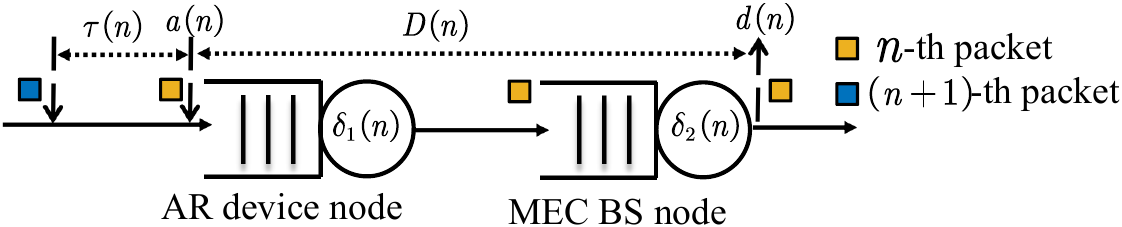}}
\caption{Tandem queueing model of the AR system.} 
\vspace{-0.2cm}
\label{Tandem queueing model}
\end{figure}

AR services have strong requirements for low E2E latency, high reliability, and high data rates. The AR system can be modeled as a tandem queuing system in the time domain, with the AR user and the BS as service nodes, as shown in Fig.~\ref{Tandem queueing model}. In the UL, the live video is segmented into frames, and the bits in a frame are represented as a packet. The average inter-arrival time between packets is the inverse of the frame rate $f$. The instantaneous inter-arrival time $\tau\left(n\right)$ between the $n$-th packet and the $(n+1)$-th packet is random due to jitter, which follows a truncated Gaussian distribution with mean $\mu$, variance $\sigma^2$ and lies within the interval $\left(b_1, b_2\right)$ according to 3GPP~specifications~\cite{Standard}.

Given the randomness in the packet arrival process and the fluctuation of UL and DL wireless channels, packets may accumulate in queues at both the AR device and the BS. Let $a\left(n\right)$ denote the arrival time of the $n$-th packet, and $d\left(n\right)$ denote the time when the augmented video frame corresponding to the $n$-th packet is transmitted to the user. Then, the E2E latency of the $n$-th packet can be expressed as
\vspace{-0.1cm}
\begin{equation}
D\left( n \right) =d\left( n \right) -a\left( n \right) ,
\label{Definition of delay}
\end{equation}	
where $D\left( n\right)$ includes UL transmission delay, UL queuing delay, DL transmission delay, and DL queuing delay. The computing delay at the base station is negligible compared to the PDB, due to the deployment of high-performance hardware and rendering algorithms, and is thus omitted here~\cite{Prediction_communication}.

For AR services, receiving a packet late is nearly as detrimental as losing it entirely \cite{Standard}. Consequently, reliability is often characterized by the PLR, which includes both the probability of packet transmission errors and the probability that a packet's E2E latency exceeds the PDB. Owing to the strong error‑correction capability of channel coding and the employment of closed‑loop link‑adaptation techniques, packet transmission errors are effectively controlled. As a result, the PLR is primarily determined by the PDB violation probability~\cite{uVR}. Thus, the QoS related to latency, reliability, and data rates is reflected by both the PDB $D_{\max}$ and the target PLR $\varepsilon_{\max}$, which is defined as
\begin{align}
  P\left( D\left( n \right) \geq D_{\max} \right) \leq \varepsilon_{\max}. \label{E:ARqos}
\end{align}

\subsection{Transmission Model}
To ensure high data rates of the AR system, a wideband orthogonal frequency division multiplexing transmission system is employed with \(N_R^u\) resource blocks (RBs) in the UL and \(N_R^d\) RBs in the DL. The bandwidth of each RB is $W_0$, and each RB spans one slot with a duration of $T_f$. Suppose that the large-scale channels remain constant across RBs while the small-scale channels are independent and identically distributed (i.i.d.) across different RBs and remain constant within each RB. The BS adopts maximum ratio combining and maximum ratio transmission for UL and DL to enhance the signal quality, respectively. The data rates in the $t$-th slot for UL and DL are expressed as
\vspace{-0.1cm}
\begin{subequations} \label{Capacity}
    \begin{align}
    R_{t}^{u}=W_0\sum\nolimits_{i=1}^{N_{R}^{u}}{\log _2}\bigg( 1+p_{i,t}^{u}\gamma_{i,t}^{u} \bigg)   \label{UL capacity}\\
    R_{t}^{d}=W_0\sum\nolimits_{i=1}^{N_{R}^{d}}{\log _2}\bigg( 1+p_{i,t}^{d}\gamma_{i,t}^{d} \bigg)   \label{DL capacity}
    \end{align}
\end{subequations}
where $p_{i,t}^{u}$ and $p_{i,t}^{d}$ denote the transmit power allocated to the $i$-th RB in the $t$-th slot for UL and DL, respectively, $\gamma_{i,t}^{u}\triangleq\frac{\alpha g_{i,t}^{u}}{W_0N_0 + \rho^2_I}$ and $\gamma_{i,t}^{d}\triangleq\frac{\alpha g_{i,t}^{d}}{W_0N_0 + \rho^2_I}$ denote the signal-to-interference-plus-noise ratios (SINRs) with unit transmit powers, $\alpha$ is the large-scale channel gain,
$g_{i,t}^{u}=\|\mathbf{h}_{i,t}^{u}\|^2$ and $g_{i,t}^{d}=\|\mathbf{h}_{i,t}^{d}\|^2$ represent the instantaneous channel gains, $\mathbf{h}_{i,t}^{u}, \mathbf{h}_{i,t}^{d} \in \mathbb{C} ^{N_t \times 1}$ are the small-scale channel vectors, 
$N_0$ is the power spectral density of noise, and $\rho_I^2$ is the inter-cell interference power.
Assume that the inter-cell interference power follows a Bernoulli distribution with an occurrence probability of $P_I$, which remains constant within a single slot and i.i.d. across different slots. % 

Due to the large amount of data in each packet, transmitting a packet typically requires multiple time slots. The total transmission time, referred to as service time, for the $n$-th packet in UL and DL is denoted as $\delta_u\left( n \right)$ and $\delta_d\left( n \right)$, respectively, which can be expressed as
%\vspace{-0.1cm}
\begin{subequations} \label{Service time}
    \begin{align}
    \!\!\!\delta _u\left( n \right) = \min \left\{\! k\in \mathbb{N}_+\!\left| \sum\nolimits_{t=t_{n}^{u}}^{t_{n}^{u}+k-1}{R_{t}^{u}\!\cdot\! T_f}\ge M_u \right. \right\}\! \cdot\! T_f
    \label{UL service time}\\
    \!\!\!\delta_d\left( n \right) = \min \left\{\! k\in \mathbb{N}_+\!\left| \sum\nolimits_{t=t_{n}^{d}}^{t_{n}^{d}+k-1}{R_{t}^{d}\!\cdot\! T_f}\ge M_d \right. \right\}\! \cdot\! T_f\label{DL service time}
    \end{align}
\end{subequations}
where $M_u$ and $M_d$ represent the sizes of the packets transmitted in the UL and DL, respectively, and $t_{n}^{u}$ and $t_{n}^{d}$ denote the indices of the slots when the transmission of the $n$-th packet starts in the UL and DL. 

\section{Problem Formulation}
To save the bidirectional transmit power, we optimize the power allocation to minimize a utility function that increases with the average UL and DL transmit power, denoted by $U(E_{t}\left[ \mathbf{1}^T\mathbf{P}_{t}^{u} \right], E_{t}\left[ \mathbf{1}^T \mathbf{P}_{t}^{d}\right])$, where $\mathbf{P}_{t}^{u}=\big[ p_{1,t}^{u},\dots,p_{N_{R}^{u},t}^{u} \big]^T$ and $\mathbf{P}_{t}^{d}=\big[ p_{1,t}^{d},\dots,p_{N_{R}^{d},t}^{d} \big]^T$. The optimization problem under QoS constraints is formulated as
%\vspace{-0.1cm}
\begin{subequations} \label{P0:Power0}
    \begin{align}
\mathrm{P}0:\min_{\mathbf{P}_{t}^{u}, \mathbf{P}_{t}^{d}} \ &U(E_{t}\left[ \mathbf{1}^T\mathbf{P}_{t}^{u} \right], E_{t}\left[ \mathbf{1}^T \mathbf{P}_{t}^{d}\right])  \label{P0:object0}\\
    \mathrm{s}.\mathrm{t}.\
    &p_{i,t}^{u}\geq 0, i=1,...,N_{R}^{u}, \forall t\in \mathbb{N} ^+, \label{P0:resctionUL0}\\
    &p_{i,t}^{d}\geq 0, i=1,...,N_{R}^{d}, \forall t\in \mathbb{N} ^+, \label{P0:resctionDL0}\\
    &\eqref{Definition of delay},\eqref{E:ARqos},\eqref{Capacity},\eqref{Service time}. \nonumber 
    \end{align}
\end{subequations}
where as indicated by the E2E latency definined in \eqref{Definition of delay}, the QoS constraint in \eqref{E:ARqos} hinges on the service times \eqref{Service time}, which themselves depend on power allocation in \eqref{Capacity}.

Directly solving $\mathrm{P}0$ is intractable because the probabilistic constraint \eqref{E:ARqos}, coupled with the service process and power allocation, lacks an explicit expression. To obtain a tractable formulation, we first derive an upper bound on the PDB violation probability, identify the conditions on the service process that ensure this bound, and then optimize the power allocation subject to these conditions.

\subsection{PDB Violation Probability for AR System}
The AR system consists of two cascaded 
queues, as shown in Fig.~\ref{Tandem queueing model}. We first examine the case with a single queue before addressing the case involving cascaded queues.

For each queue, either the UL or the DL queue, the packet delay of the $n$-th packet, as defined in \eqref{Definition of delay}, can be derived using the max-plus queueing principle~\cite{Two_side} as
\begin{equation} \label{Queueing principle10}
\begin{split}
    D\left( n \right) %&=a\overline{\otimes }\Delta (0,n)-a(n) \nonumber\\
    &=\max_{0\le m\le n} \left[\Delta \left( m,n \right) - \Gamma \left(m,n \right) \right]\\  %\label{Queueing principle1} \\
    &=\max_{0\le m\le n} \left[ \sum\nolimits_{k=m}^{n}{\delta}(k)-\sum\nolimits_{k=m}^{n-1}{\tau}(k) \right], %\label{Queueing principle2}
\end{split}
\end{equation}
where $\Delta \left( m,n \right) =\sum_{k=m}^n{\delta \left( k \right)}$ and $\Gamma \left( m,n \right) =\sum_{k=m}^{n-1}{\tau \left( k \right)}$ represent the cumulative service time from the $m$-th packet to the $n$-th packet and the inter-arrival time between the $m$-th packet and the $n$-th packet, respectively, and $\delta=\delta_u$ or $\delta_d$ for the UL or DL queue. With \eqref{Queueing principle10}, the PDB violation probability is derived as
\begin{equation} \label{E:pdb_prob}
\begin{split}
&P\{D(n)>D_{\max}\} \\
&=P\left\{ e^{\theta \underset{0\leq m\leq n-1}{\max}\sum\nolimits_{k=m}^{n-1}{\left[ \delta \left( k \right) -\tau \left( k \right) +\delta \left( n \right) \right]}}>e^{\theta D_{\max}} \right\}   \\
&=P\left\{ \max_{0\le l\le n-1} V\left( l \right) >e^{\theta D_{\max}} \right\},
\end{split}
\end{equation}
where $V\left( l \right) \triangleq e^{\theta \sum\nolimits_{k=n-1-l}^{n-1}{\left[ \delta \left( k \right) -\tau \left( k \right) +\delta \left( n \right) \right]}}$, and $\theta>0$.

Since both the inter-arrival time $\tau(k)$  and the service time $\delta(k)$ are i.i.d., it can be proved that the sequence $\left\{ V\left( l \right),\ l=0, 1, \cdots, n-1 \right\} $ forms a supermartingale for all $\theta>0$ satisfying~\cite{Two_side}
\vspace{-0.1cm}
\begin{equation}\label{E:delay001}
  E\left[ e^{\theta \delta \left( n \right)} \right] E\left[ e^{-\theta \tau \left( n \right)} \right] \leq 1.
\end{equation}
Under \eqref{E:delay001}, applying Doob’s inequality to \eqref{E:pdb_prob} yields the following upper bound~\cite{Two_side}
\begin{align}
&P\{D(n)>D_{\max}\} \leq \frac{1}{E\left[ e^{-\theta \tau (n)} \right]}e^{-\theta D_{\max}}. \label{Delay bound23}
\end{align}

Next, we consider the tandem queueing system with two service nodes. The results presented below can be generalized to systems with more than two nodes. By extending \eqref{Queueing principle10} as 
\vspace{-1ex}
\begin{equation} \label{Queueing principle11}
\begin{split}
    &D\left( n \right) 
    =\max_{0\le m_1\le n} \left[\Delta_1\overline{\otimes }\Delta _2 \left( m_1,n \right) - \Gamma \left(m_1,n \right) \right]\\  %\label{Queueing principle1} \\
    &=\underset{0\leq m_1\leq m_2\leq n}{\max} \left[ \sum_{k=m_1}^{m_2}{\delta_1}(k)+\sum_{k=m_2}^{n}{\delta_2}(k)-\sum_{k=m_1}^{n-1}{\tau}(k) \right],\raisetag{3.5\baselineskip} %\label{Queueing principle2}
\end{split}
\end{equation}
the PDB violation probability can be expressed as
\vspace{-1ex}
\begin{equation} \label{Delay bound02}
\begin{split}
& P\left\{ D(n)> D_{\max} \right\}
\\
&= 1-P\left\{ \underset{0\leq m_1\leq m_2\leq n}{\max}\left[ \sum\nolimits_{k=m_1}^{m_2}{\delta_1\left( k \right)}-\sum\nolimits_{k=m_1}^{m_2-1}{\tau \left( k \right)} \right. \right.
\\
& \left. \left. +\sum\nolimits_{k=m_2}^n{\delta _2\left( k \right)} -\sum\nolimits_{k=m_2}^{n-1}{\tau \left( k \right)} \right] \leq D_{\max} \right\},\raisetag{1.0\baselineskip}
\end{split}
\end{equation}
where $\overline{\otimes }$ is the max-plus convolution operator, and $\delta_h\left( k \right)$ and $\Delta_h \left( m,n \right) =\sum_{k=m}^n{\delta_h \left( k \right)}$ represents the $k$-th packet service time and cumulative service time of the $h$-th node, respectively, $h=1,2 $. By relaxing the constraint $0\leq m_1\leq m_2\leq n$ in \eqref{Delay bound02} to two independent constraints $0\leq m_1\leq n$ and $0\leq m_2\leq n$, the feasible range of $m_1$ and $m_2$ is expanded, leading the following inequality
\vspace{-1ex}
\begin{equation}\label{Delay bound2}
\begin{split}
& P\left\{ D(n)> D_{\max} \right\}
\\
&\leq 1-P\left\{ \underset{0\leq m_1\leq n}{\max}\left[ \sum\nolimits_{k=m_1}^n{\delta_1\left( k \right)}-\sum\nolimits_{k=m_1}^{n-1}{\tau \left( k \right)} \right] \right.
\\
&\left.+\underset{0\leq m_2\leq n}{\max}\left[ \sum\nolimits_{k=m_2}^n{\delta _2\left( k \right)}-\sum\nolimits_{k=m_2}^{n-1}{\tau \left( k \right)} \right] \leq D_{\max} \right\}.\raisetag{4.5\baselineskip}
\end{split}
\end{equation}

Since the small-scale channels are i.i.d. across different RBs, it is reasonable to assume the independence between the service times $\delta_1\left(k\right)$ and $\delta _2\left(k\right)$.  Thus the two max-plus terms $\underset{0\leq m_1\leq n}{\max}\big[ \sum\nolimits_{k=m_1}^n{\!\delta_1\left( k \right)}\!-\!\sum\nolimits_{k=m_1}^{n-1}{\!\tau \left( k \right)} \big]$ and $\underset{0\leq m_2\leq n}{\max}\big[ \sum\nolimits_{k=m_2}^n{\!\delta _2\left( k \right)}\!-\!\sum\nolimits_{k=m_2}^{n-1}{\!\tau \left( k \right)} \big]$ are statistically independent for any given distribution of $\tau \left( k \right)$. As a result of this independence, and noting that both max-plus terms share the same form as \( D(n) \) in \eqref{Queueing principle10}, whose complementary cumulative distribution function is derived in \eqref{Delay bound23}, the Stieltjes convolution of their distribution functions yields the following bound~\cite{SNC_jiang}
\vspace{-1ex}
\begin{equation}
P\left\{ D(n) > D_{\max} \right\} \leq 1 - \overline{f}_1 \ast \overline{f}_2 \left( D_{\max} \right),
\label{Delay bound24}
\end{equation}
where \(\ast\) denotes the Stieltjes convolution operator and $\overline{f}_h = 1 - \min\{f_h(\theta,x),1\}$, with $f_h(\theta,x) = \frac{1}{E \left[ e^{-\theta\tau(n)} \right]} e^{-\theta x}$ denoting the upper bound on the PDB violation probability for the $h$-th queue as given in \eqref{Delay bound23}. \(\theta > 0\) is the QoS exponent characterizing the exponential decay rate of the violation probability, which satisfies \(\mathbb{E}[e^{\theta \delta_h(n)}] \, \mathbb{E}[e^{-\theta \tau(n)}] \leq 1\) for \(h = 1, 2\).

Based on \eqref{Delay bound24}, the PDB violation probability of the AR system is upper bounded by
\vspace{-1ex}
\begin{align}
  P\left( D\left( n \right) \geq D_{\max} \right) \leq 1-\overline{f}_u\ast \overline{f}_d\left(\theta, D_{\max} \right), \label{E:qosUB}
\end{align}
where $\overline{f}_u$ and $\overline{f}_d$ are defined analogously to $\overline{f}_1$ and $\overline{f}_2$ in \eqref{Delay bound24}. As indicated by \eqref{E:delay001}, the validity of this upper bound requires that, for a given $\theta>0$, the service times in \eqref{Service time} satisfy the following conditions
\vspace{-1ex}
\begin{subequations} \label{Resction}
  \allowdisplaybreaks
  \begin{align}
    &E\left[ e^{\theta\delta_u(n)} \right] E\left[ e^{-\theta\tau(n)} \right] \leq 1, \label{UL Resction} \\
    &E\left[ e^{\theta\delta_d(n)} \right] E\left[ e^{-\theta\tau(n)} \right] \leq 1. \label{DL Resction}
  \end{align}
\end{subequations}

With \eqref{E:qosUB}, the QoS requirement of the AR system given in \eqref{E:ARqos} is relaxed to the constraint on the upper bound of PDB violation probability, which is
\vspace{-1ex}
\begin{align}
  1 - \overline{f}_u\ast \overline{f}_d\left(\theta, D_{\max} \right) \leq \varepsilon_{\max}. \label{E:qosUB1}
\end{align}

\vspace{-1.0ex}
\subsection{Approximation of Service Time Distribution}
The conditions in \eqref{Resction} involve the terms \( E\left[ e^{\theta \delta_u(n)} \right] \) and \( E\left[ e^{\theta \delta_d(n)} \right] \), where the service times \(\delta_u(n)\) and \(\delta_d(n)\) are functions of the power allocation. However, as indicated in \eqref{Service time}, the distributions of \(\delta_u(n)\) and \(\delta_d(n)\) lack closed-form expressions, making direct derivation intractable. To address this, we approximate the distributions of \(R_t^u\) and \(R_t^d\) as a Gaussian distribution using the central limit theorem (CLT), thereby facilitating tractable approximations of \(\delta_u(n)\) and \(\delta_d(n)\). Since the computations of \( E\left[ e^{\theta \delta_u(n)} \right] \) and \( E\left[ e^{\theta \delta_d(n)} \right] \) follow similar procedures, we use the UL case to illustrate the Gaussian approximation and subsequent derivation of \( E\left[ e^{\theta \delta_u(n)} \right] \).

Specifically, as described in \eqref{UL capacity}, $R_t^u$ is defined as the sum of data rates over \(N_R^u\) RBs. Considering that small-scale fading channels across RBs are i.i.d. and \(N_R^u\) is large, the CLT implies that $R_t^u$ can be approximated by two Gaussian distributions which are $R_I\sim N\left( \mu _I, \sigma _{I}^{2} \right)$ and $R_U\sim N\left( \mu _U, \sigma _{U}^{2} \right)$, corresponding to scenarios with and without inter-cell interference, respectively. The statistical parameters $\mu _I$, $\sigma _{I}^{2}$, $\mu_U$, and $\sigma _{U}^{2}$ can be computed as 
\begin{subequations} \label{tatistical parameters}
    \begin{align}
    &\mu _I=\frac{1}{\left| \mathcal{B} _I \right|}\sum_{i\in \mathcal{B} _I}{R_{i}^{u}}, &\sigma _{I}^{2}=\frac{1}{\left| \mathcal{B} _I \right|}\sum_{i\in \mathcal{B} _I}{\left( R_{i}^{u}-\mu _I \right) ^2}
\\
&\mu _U=\frac{1}{\left| \mathcal{B} _U \right|}\sum_{i\in \mathcal{B} _U}{R_{i}^{u}}, &\sigma _{U}^{2}=\frac{1}{\left| \mathcal{B} _U \right|}\sum_{i\in \mathcal{B} _U}{\left( R_{i}^{u}-\mu _U \right) ^2}
    \end{align}
\end{subequations}
where \( R_i^u \) is defined in \eqref{UL capacity}, \( \mathcal{B}_I \) and \( \mathcal{B}_U \) denote the sets of samples collected under conditions with and without inter-cell interference, respectively, and \( |\mathcal{B}_I| \) and \( |\mathcal{B}_U| \) represent the number of samples.

Given \(R_I\) and \(R_U\), the distribution of the cumulative service rate over $k$ slots can be analyzed as follows. Consider that there are $i$ slots out of the $k$ slots experiencing inter-cell interference with probability $P_{k,i}=\binom{k}{i}P_I^{\,i}(1-P_I)^{k-i}$. Under this condition, the cumulative service rate can be expressed as $R_{k,i,t}=\sum\nolimits_{t}^{t+k-1}{R_{t}^u}$, where $i$ terms following \(R_I\), $(k-i)$ terms following \(R_U\), and $t$ denotes the index of the starting time slot. Consequently, $R_{k,i,t}$ follows a Gaussian distribution with mean \(\mu_{i,k}=i\mu_I+(k-i)\mu_U\) and standard deviation \(\sigma_{i,k}=\sqrt{i\sigma_I^{2}+(k-i)\sigma_U^{2}}\). And, the probability that the service time of the \(n\)-th packet does not exceed \(k\) slots can be derived as
%\vspace{-0.15cm}
\begin{equation}\label{Service time CDF}
\begin{split}
&P\left\{ \delta_u \left( n \right) \leq k\cdot T_f \right\} =\sum_{i=0}^kP_{k,i}\cdot P\left\{{R_{k,i,t_n^u}^u\cdot T_f}\geq M_u|i \right\} \\
&=\sum_{i=0}^k{\left(P_{k,i}\cdot Q\left( \frac{M_u-\mu _{i,k}}{\sigma _{i,k}} \right)\right)}\triangleq \Pr\left( M_u,k \right), \raisetag{1.5\baselineskip}
\end{split}
\end{equation}
where $P\left\{{R_{k,i,t_n^u}^u\cdot T_f}\geq M_u|i \right\}$ denotes the conditional probability given $i$ interfering slots, \(Q(\cdot)\) is the Gaussian \(Q\)-function, and $k\in \mathbb{N}_+$. 

Then, the distribution of $\delta_u$ can be obtained as
%\vspace{-0.1cm}
\begin{equation} \label{Service time PDF}
\!\!\!P\left\{\! \delta_u \left( n \right)\!=\!k T_f \!\right\}\!=\!\begin{cases}
	\!\Pr\left(\! M_u,k \!\right), &k=1,\\
	\!\Pr\left(\! M_u,k\! \right)\!-\!\Pr\left(\! M_u,k-1 \!\right),\!\!\!\!\!  &k>1.
\end{cases}
\end{equation}
With \eqref{Service time PDF}, the expectation $E\left[ e^{\theta\delta_u \left( n \right)} \right] $ in constraint \eqref{UL Resction} can be computed as
\vspace{-0.1cm}
\begin{align} \label{MGF}
E\left[ e^{\theta\delta_u \left( n \right)} \right] =\sum\nolimits_{k=1}^{K_{\max}}{{P}\left\{ \delta_u \left( n \right) =kT_f \right\} e^{\theta kT_f}},
\end{align}
where $K_{\max}$ is the number of slots leading to negligible probability ${P}\left\{ \delta_u \left( n \right) =K_{\max}T_f \right\}$.

\subsection{Power Optimization Problem}
Since the QoS exponent $\theta$ in constraints \eqref{Resction} and \eqref{E:qosUB1} influences the tightness of the upper bound and service process, and consequently the power allocation, we first determine its optimal value. Noting that increased PDB violation probability is associated with lower transmit power, the optimal $\theta^\star$ is achieved when \eqref{E:qosUB1} holds with equality, i.e.,
%\vspace{-0.1cm}
\begin{align}
  1 - \overline{f}_u\ast \overline{f}_d\left(\theta, D_{\max} \right) = \varepsilon_{\max}. \label{E:qosUB2}
\end{align}
Since $\overline{f}_u$ and $\overline{f}_d$ depend solely on the arrival process, as defined below \eqref{Delay bound24}, the optimal value $\theta^\star$ can be readily obtained, e.g., by a bisection method with the arrival process~parameters.

% \addtolength{\topmargin}{0.01in}
By setting $\theta = \theta^\star$, and replacing probabilistic constraints \eqref{Definition of delay} and \eqref{E:ARqos} in $\mathrm{P}0$ with \eqref{Resction}, we find that constraints in $\mathrm{P}0$ can be decoupled into UL and DL constraints. Furthermore, since the utility function increases with the average UL and DL transmit power, problem \eqref{P0:Power0} can be decomposed into two subproblems: one optimizing the average UL power and the other optimizing the average DL power. In the sequel, we focus on the power allocation of one link, while the power allocation for the other link can be derived in the same way. Taking UL as an example, the subproblem can be finally formulated~as
\begin{subequations} \label{P2:Power}
    \begin{align}
    \mathrm{P}1: & \min_{\mathbf{P}_{t}^{u}}\ E_{t}\left[ \mathbf{1}^T\mathbf{P}_{t}^{u} \right]  \label{P2:object}\\
    &\mathrm{s}.\mathrm{t}.\
    p_{i,t}^{u}\geq 0, i=1,...,N_{R}^{u}, \forall t\in \mathbb{N} ^+, \label{P2:resctionUL}\\
    & \eqref{UL capacity},\eqref{UL service time},\eqref{UL Resction},\eqref{Service time CDF}, \eqref{Service time PDF},\eqref{MGF},\eqref{E:qosUB2}\nonumber.
    \end{align}
\end{subequations}
where the objective function is the average UL transmit power considering the monotonicity of  $U(E_{t}\left[ \mathbf{1}^T\mathbf{P}_{t}^{u} \right])$.

\section{Learning Power Allocation with DNN}
The objective of problem $\mathrm{P}1$ is to minimize the average power across slots, with instantaneous power allocation \(\mathbf{P}_{t}^{u}\) adapting to the instantaneous SINR \(\boldsymbol{\gamma}_{t}^u = \left[\gamma_{1,t}^u, \cdots, \gamma_{N_R^u,t}^u\right] \in \mathbb{R}^{N_R^u \times 1}\). This forms a functional optimization problem, intended to optimize the mapping from $\boldsymbol{\gamma}_{t}^u$ to $\mathbf{P}_{t}^{u}$, denoted as \(\mathbf{P}_{t}^{u} = \phi(\boldsymbol{\gamma}_{t}^u)\). Solving this problem is challenging because constraint \eqref{UL Resction} depends on the statistic parameters $\mu _I$, $\sigma _{I}^{2}$, $\mu_U$, and $\sigma _{U}^{2}$, which are related to the power allocation $\mathbf{P}_{t}^{u}$ but lack an explicit expression for their relationship. Therefore, we resort to deep learning to tackle this problem.

According to the proof in \cite{Probabilistic_constrained}, a functional optimization problem with constraints can be solved by unsupervised primal-dual learning. Specifically, for problem $\mathrm{P}1$, the power allocation policy $\phi(\boldsymbol{\gamma }_{t}^u)$ can be parameterized by the DNN ${\Phi}\left( \boldsymbol{\gamma }_{t}^u;\boldsymbol{\theta }_{\mathbf{P}} \right)$, where $\boldsymbol{\theta }_{\mathbf{P}}$ denotes the trainable parameters. The primal-dual problem can be expressed as
\begin{subequations} \label{P4:primal-dual problem}
    \begin{align}
    &\underset{\lambda}{\max}\,\,\underset{\boldsymbol{\theta }_{\mathbf{P}}}{\min}\,\,\mathcal{L} \left( \Phi\left( \boldsymbol{\gamma }_{t}^u;\boldsymbol{\theta }_{\mathbf{P}} \right) ,\lambda \right) \,\, \label{P4:object}\\
     &\ \ \mathrm{s}.\mathrm{t}. \ \ \  \Phi\left( \boldsymbol{\gamma }_{t}^u; \boldsymbol{\theta }_{\mathbf{P}} \right)  \succeq 0, \ \ \lambda \succeq 0, \label{P4:resction1}
    \end{align}
\end{subequations}
where $\mathcal{L} \left( \Phi\left( \boldsymbol{\gamma }_{t}^u;\boldsymbol{\theta }_{\mathbf{P}} \right),\lambda\right)$ is the Lagrangian function,
\begin{equation}  \label{P4:lagrange}
\begin{split}
  &\mathcal{L} \left(\Phi\left( \boldsymbol{\gamma }_{t}^u;\boldsymbol{\theta }_{\mathbf{P}} \right),\lambda\right) = E_{t}\left[ \mathbf{1}^T\Phi\left( \boldsymbol{\gamma }_{t}^u;\boldsymbol{\theta }_{\mathbf{P}} \right)  \right] + \\
  &\  \lambda \left(\!\sum_{k=1}^{K_{\max}}{{P}\left\{ \delta_u \left( n \right) \!=\! kT_f \right\} e^{\theta^\star kT_f}} \!-\! \frac{1}{E\left[ e^{-\theta^\star\tau \left(n\right)} \right]} \right),
\end{split}
\end{equation}
and $\lambda$ is the Lagrangian multiplier. The parameters $\boldsymbol{\theta }_{\mathbf{P}}$ and the multiplier $\lambda$ can be iteratively updated along the descent and ascent directions of the sample-averaged gradients of the Lagrangian function $\mathcal{L} \left(\Phi\left( \boldsymbol{\gamma }_{t}^u;\boldsymbol{\theta }_{\mathbf{P}} \right),\lambda\right)$, respectively. 

In what follows, we design the DNN ${\Phi}\left( \boldsymbol{\gamma }_{t}^u;\boldsymbol{\theta }_{\mathbf{P}} \right)$. The power allocation policy $\mathbf{P}_{t}^{u} = \phi(\boldsymbol{\gamma }_{t}^u)$ satisfies the permutation equivariance (PE) property, which can be expressed as 
\begin{equation} \label{E:PE}
  \mathbf{\Pi}\mathbf{P}_{t}^{u} = \phi(\mathbf{\Pi}\boldsymbol{\gamma }_{t}^u),
\end{equation}
where $\mathbf{\Pi}$ is an arbitrary permutation matrix. This suggests the need to design a DNN satisfying the PE property.

\begin{figure}%[htbp]
%\vspace{-0.1cm}
\centerline{\includegraphics[width=0.30\textwidth]{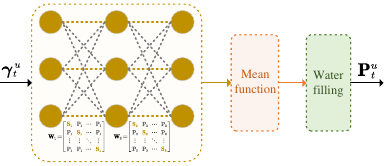}}
\caption{Illustration of the DNN architecture.} \label{GNN}
\vspace{-0.5cm}
\end{figure}

Instead of directly learning the power allocation for each RB, we exploit the optimal structure of the power allocation. We can prove that the optimal power allocation has a water filling structure by contradiction. First, suppose the optimal policy does not follow a water-filling approach. For any given total power allocated by such an optimal policy within a slot, a water-filling policy can always achieve a higher data rate. Thus, it would be possible to further reduce the total power while still satisfying the QoS requirements, contradicting the assumption of optimality. Given the water-filling structure, the learning task can be simplified to learn the scalar water-level, which can significantly improve learning efficiency.

The DNN architecture for learning the power allocation policy is shown in Fig.~\ref{GNN}. Specifically, the first module is a DNN with a parameter-sharing weight matrix $\mathbf{W}$ designed as in \cite{Deep_models} to ensure the PE property is satisfied, the second module employs a mean function to determine the water-level scalar, and the third module implements the fixed water-filling procedure. It can be easily verified that the proposed DNN satisfies the PE property in \eqref{E:PE}.

\section{Simulation Results} \label{Simulation}
In this section, we evaluate the power allocation performance of the proposed DNN in the AR system by comparing it with the relevant baselines.

\subsection{Simulation Setup} 
The simulation parameters are summarized in Table~\ref{parameter}. The DNN, depicted in Fig.~\ref{GNN}, is trained through an unsupervised learning approach, employing the Lagrangian function defined in \eqref{P4:lagrange} as the loss function. The channels follow the Rayleigh fading model, and we generate 500,000 channel samples for training and an additional 10,000 samples for testing. The hyperparameters used in the simulation are fine-tuned as follows. The dimensions of the hidden layers are set to [1, 256, 512, 512, 256, 1], with Tanh activation functions applied to each hidden layer and a ReLU applied to the output layer. The batch size is set to be 1024, the Adam optimizer is employed, and the initial learning rate is 0.001.

\begin{table}% [htbp]
%\vspace{-0.1cm}
\centering
\caption{SIMULATION PARAMETERS AND HYPER-PARAMETERS}
\begin{tabular}{c|c}
\hline
\hline
\text{Target PLR} $\varepsilon_{\max}$  & $10^{-3}$ \\
\hline
\text{PDB} $D_{\max}$  &  20 ms \\
\hline
\text{Duration of each slot} $T_f$ & 1 ms \\
\hline
\text{Frame size} $M_u$, $M_d$ & $0.1$, $1$ Mb  \\
\hline
\text{Frame per second} $f$ & 120 fps  \\
\hline
\multirow{2}{*}{Truncated Gaussian distributed}  & $\left(\mu, \sigma, b_1, b_2\right)$\\
arrival process & $(\frac{1}{f}, 2, \frac{1}{f}-5, \frac{1}{f}+5)$\\
\hline
\text{Path loss model} $10\log\alpha$  & 35.3 + 37.6$\log$100 \\
\hline
\text{Number of antennas} $N_{t}$  & 8 \\
\hline
\text{Single-sided noise spectral density} $N_{0}$  & -173 dBm/Hz \\
\hline
\text{Number of RBs} $N_{R}^u$, $N_{R}^d$ & 52, 133 (10MHz, 20MHz BW)\\
\hline
\text{Bandwidth of RBs} $W_0$  & 180 kHz \\
\hline
\text{Interference probability} $P_I$  & 0.5 \\
\hline
\text{INR} $\eta=10 \log_{10}(\frac{\rho_I^2}{W_0N_0}) $  & 0, 5, 10, 15, 20 dB \\
\hline
\hline
\end{tabular}
\label{parameter}
\vspace{-0.5cm}
\end{table}

\subsection{Simulation Performance} 
\begin{figure}%[htbp]
\centering
\includegraphics[width=0.40\textwidth]{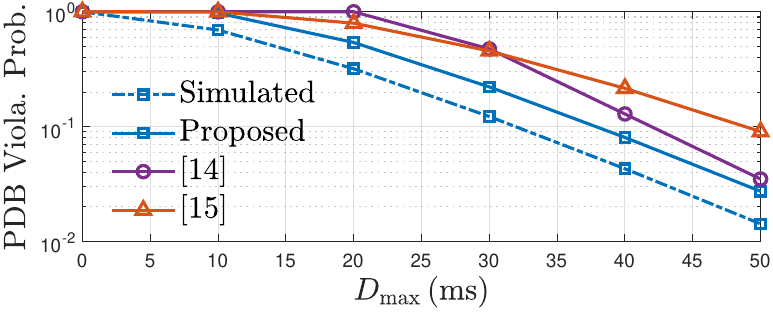}
\caption{Tightness of bounds for PDB violation probability. }  \label{tightness}
\end{figure}

We first evaluate the tightness of the derived upper bound of PDB violation probability by comparing it with existing upper bounds derived in \cite{A_combined, Dealing_wit}, where the inter-arrival times of data packets follow a truncated Gaussian distribution, as specified in Table~\ref{parameter} and the service times for the two nodes follow an exponential distribution with an average service time of 5 slots. Fig.~\ref{tightness} shows that the derived upper bound closely approximates the simulated results, demonstrating the tightness of the proposed bound.

Then, to evaluate the accuracy of the Gaussian approximation, a Kolmogorov-Smirnov test is conducted. The test provides a p-value that indicates the likelihood that the observed sample rates follow a Gaussian distribution. We obtain a p-value of 0.8335 for the sample rates, suggesting the accuracy of the Gaussian approximation.

\begin{figure}%[htbp]
%\vspace{-0.1cm}
\centering
\subfigure[UL average power saving gain]{
\includegraphics[width=0.21\textwidth]{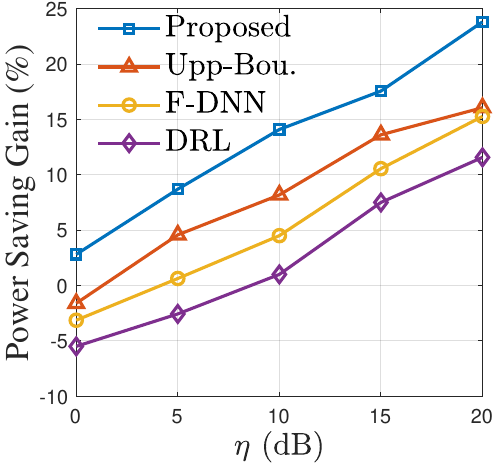} \label{UL Power}
}
\hfill
\subfigure[DL average power saving gain]{
\includegraphics[width=0.21\textwidth]{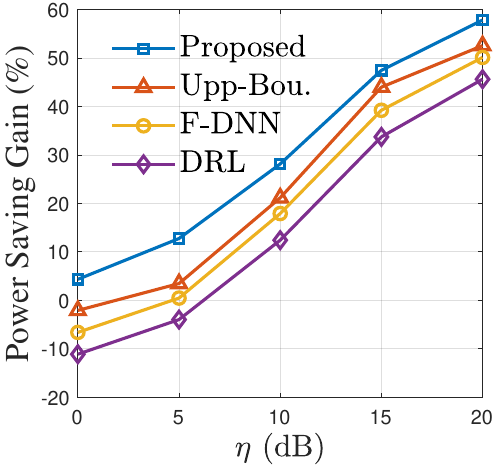} \label{DL Power}
}
\caption{Performance comparison at different INRs. } \label{Power perforance}
%\vspace{-0.5cm}
\end{figure}

Finally, the performance of the proposed method is compared with the following policies. 
\begin{itemize}
	\item[$\bullet$] \textbf{Non-DNN:} As assumed in \cite{Asynchronous_hybrid}, each packet has an equal and constant UL and DL service time, determined by \eqref{Resction}. We further assume that bits within each packet are uniformly transmitted across the available slots. Moreover, the optimal water-filling algorithm is applied in each slot to minimize the total transmit power.
    
    \item[$\bullet$] \textbf{Upp-Bou.:} This method utilizes the theoretical bound derived in \cite{Dealing_wit} to calculate the value of $\theta^\star$, and employs the proposed learning method for power allocation.

    \item[$\bullet$] \textbf{F-DNN:} In this approach, $\theta^\star$ is determined by the upper bound derived in \eqref{E:qosUB2}. A fully connected neural network, as described in \cite{Probabilistic_constrained}, is then used for power allocation.

    \item[$\bullet$] \textbf{DRL:} Similar to F-DNN, $\theta^\star$ is calculated based on the upper bound \eqref{E:qosUB2}. Power allocation is optimized through deep reinforcement learning, where the states, actions, and reward functions are designed following~\cite{AoI_violation_probability}.
    %\vspace{-0.5cm}
\end{itemize}.

Fig.~\ref{UL Power} and Fig.~\ref{DL Power} show the power-saving gains achieved by different methods compared to Non-DNN for UL and DL, respectively, under different interference-to-noise ratios (INRs). It is observed that the proposed method achieves the highest power saving gains in both UL and DL scenarios, especially under high interference conditions. Compared with Non-DNN, the proposed method effectively exploits the tolerance of AR service to a small PDB violation probability by adapting the service processes to the arrival process. The proposed policy also outperforms Upp-Bou., owing to the tighter bound on the PDB violation probability derived in \eqref{Delay bound24}. Furthermore, incorporating the water-filling structure and PE property into the DNN design significantly enhances the learning performance, thereby surpassing other learning-based~counterparts. 

Since the proposed method integrates prior knowledge, including mathematical modeling and PE property, it exhibits reduced training complexity concerning space requirements, sample efficiency, and training time compared to alternative learning-based methods.

\section{Conclusions}
This paper studied the UL and DL power allocation for the AR system under the QoS constraints. We derived an upper bound for the violation probability of the packet delay budget and developed approximated distributions of the service times for wideband transmission. A DNN was designed to learn the power allocation policy, leveraging the structure of the optimal power allocation and the PE property to enhance learning performance. Simulation results verified the tightness of the proposed upper bound and showed the gains of the proposed method in power saving compared to baseline methods.
\begin{comment}
\end{comment}

\bibliographystyle{IEEEtran}
\bibliography{my_ref}
\end{document}